\documentclass[11pt,a4paper]{article}
\usepackage[hyperref]{acl2020}
\usepackage{times}
\usepackage{latexsym}
\usepackage{graphicx}

\usepackage{microtype}
\usepackage{booktabs}
\usepackage{amsmath}

\aclfinalcopy 

\title{CALM: Continuous Adaptive Learning for Language Modeling}

\author{Kristjan Arumae \\
  Amazon \\
  Seattle, USA \\
  \texttt{arumae@amazon.com} \\\And
  Parminder Bhatia \\
  Amazon \\
  Seattle, USA \\
  \texttt{parmib@amazon.com} \\}

\date{}

\begin{document}
\maketitle
\begin{abstract}

Training large language representation models has become a standard in the natural language processing community.
This allows for fine tuning on any number of specific tasks, however, these large high capacity models can continue to train on domain specific unlabeled data to make initialization even more robust for supervised tasks.
We demonstrate that in practice these pre-trained models present performance deterioration in the form of \textit{catastrophic forgetting} when evaluated on tasks from a general domain such as GLUE.
In this work we propose \textbf{CALM}, \textbf{C}ontinuous \textbf{A}daptive Learning for \textbf{L}anguage \textbf{M}odeling: techniques to render models which retain knowledge across multiple domains.
With these methods, we are able to reduce the performance gap across supervised tasks introduced by task specific models which we demonstrate using a continual learning setting in biomedical and clinical domains.

\end{abstract}

\section{Introduction}

Transformer \citep{NIPS2017_7181} based language representation 
has replaced many previous pre-training or initialization approaches \citep{devlin2018bert, radford2019language, yang2019xlnet, liu2019roberta}.
Fine tuning using these architectures often yields state-of-the-art results on the order of a few hours.
The caveat to these robust models is that the initial training can be on the scale of several weeks and on many distributed GPUs which is a costly endeavour.

\begin{figure}[!h]
    \centering
    \includegraphics[scale=0.055]{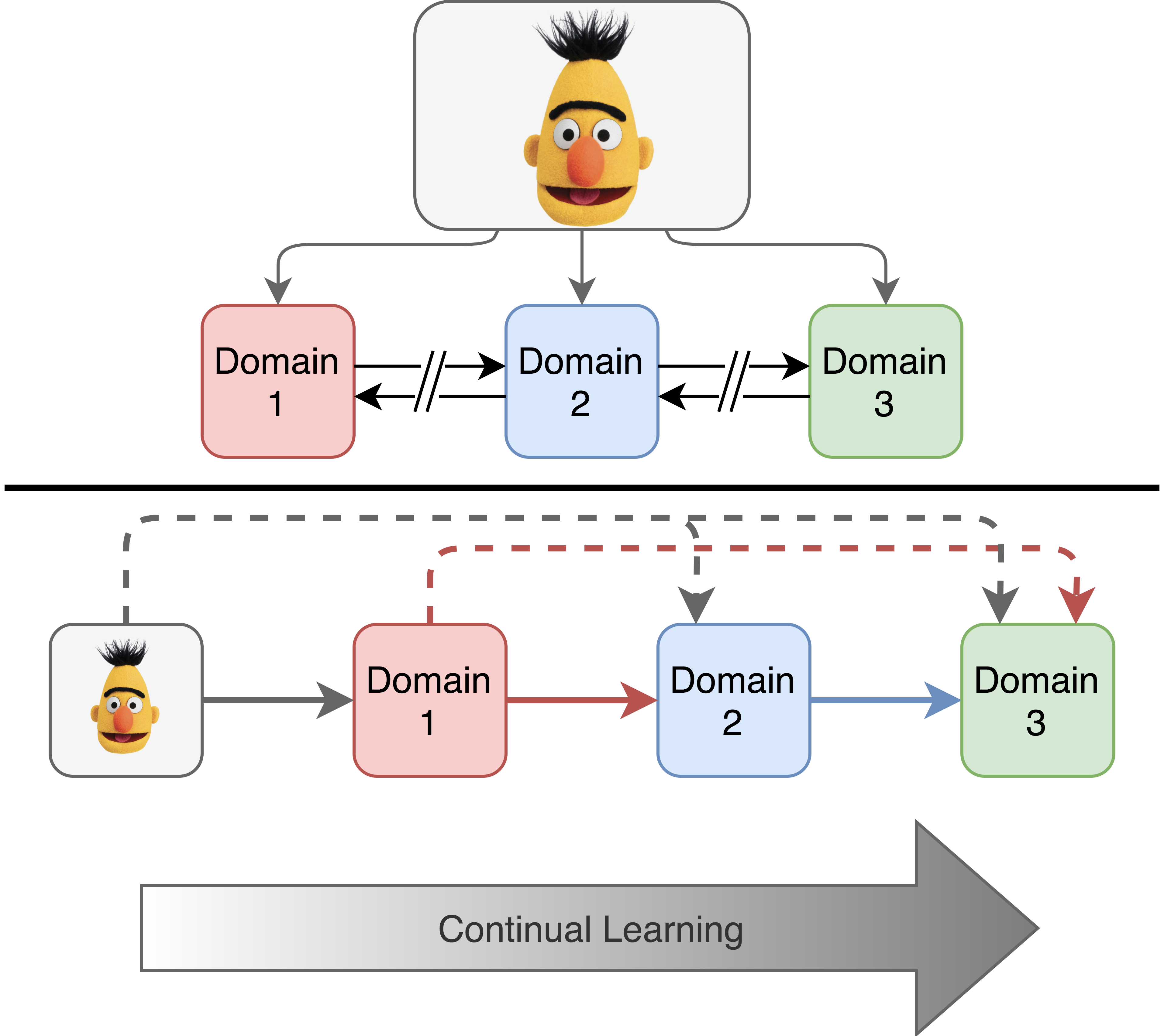}
    \centering
    \caption{Traditional approaches (top) train independent domain specific language models (blue, red, and green) which share no cross domain knowledge.  Our approach (bottom) illustrates how several domains are introduced in sequence, with knowledge retention (dashed line) using mitigation techniques, across all domains.}
    \label{fig:model}
\end{figure}

Pre-trained language models are further adapted to perform strongly in other domains as well.
For example, while the original BERT models \citep{devlin2018bert} were trained on English Wikipedia articles and BooksCorpus \citep{zhu2015aligning}, the same masked language modeling was continued on biomedical data.
BioBERT \citep{lee2019biobert} was trained using Pubmed abstracts and full articles and Clinical BERT \citep{alsentzer2019publicly} parameters were further refined using MIMIC-III clinical notes \citep{johnson2016mimic}.
Evidence suggest that understanding the syntactic structure of scientific literature and clinical data from pre-training boosts performance in their respective downstream tasks \cite{peng2019transfer}.
Training is done with the expectation of building robust, high capacity generalized language models can continue to absorb new domain knowledge.

Catastrophic forgetting \citep{mccloskey1989catastrophic, ratcliff1990connectionist} is the unfortunate side-effect of incorporating new domain data one after another. 
Parameters shift towards capturing the current task and if previous data is no longer available, the model will lose representation of it.
In general perplexity increases for older domains,
and models lose confidence in continual learning settings \cite{yogatama2019learning}.
For many tasks the straightforward solution is to combine datasets during training and approach this as a multi-task learning (MTL) \citep{ruder2017overview} problem.
Mixing data
has the desired effect of constraining parameters to find a space where both tasks reach close to optimal performance.

We further argue that these expensive pre-trained models are an example where MTL is not feasible in practice for several  reasons.
Time and hardware accessibility are the largest constraints for developing these models.
Access to processed training data is generally not possible \citep{radford2019language, devlin2018bert}, and exact training configurations are equally difficult to gather with results being arduous to reproduce.
Resource usage has recently been criticized from another perspective as well.
\citet{strubell2019energy} show that as deep neural architectures in the natural language community grow we increasingly trade results for carbon emissions. 

Current work in catastrophic forgetting mitigation has been limited to a few  small scale tested methods.
\citet{howard2018universal} introduced a multi stage training scheme for fine tuning LSTM based universal language models (ULMFiT).
The authors proposed that current methods, rather than data, are ineffective and focused on learning rate control across layers, as well as modifying learning rate scheduling.
A larger category of work deals with constraining model parameters to a latent space where they continue to capture previous tasks.
Initial work focused on model regularization and varying activations \cite{goodfellow2013empirical}. \citet{kirkpatrick2017overcoming} provided a more sophisticated solution constraining weights individually termed elastic weight consolidation (EWC).
We make use of both EWC and ULMFiT and provide further technical detail in this paper.
The final approach is focused on experience replay.
Using small samples of data from previous tasks coupled with local adaptation \citet{d2019episodic} demonstrate improvement in a Lifelong Learning (LL) training scheme.
\citet{chaudhry2019continual} also explore  LL by experimenting with updating the memory bank for experience replay.
Our work focuses on both of these techniques with the major difference being problem scale.
Many existing works apply these solutions on small networks whereas we experiment on architectures having several orders of magnitude more parameters.\newline
Our contributions are as follows:

\begin{itemize}
    \item
    We motivate the task by providing concrete evidence of catastrophic forgetting for language representation pre-training evaluated on the GLUE benchmark \cite{wang2018glue}.
    \item
    We provide empirical evidence of catastrophic forgetting mitigation with experience replay, learning rate control, and elastic weight consolidation, applied towards large scale language model pre-training. 
    \item
    We further demonstrate the robustness of elastic weight consolidation when pre-training under two stages of domain shift.
\end{itemize}



\begin{table*}
    \centering
    \begin{tabular}{l|rrrrrrrrr}
    Model & CoLA & SST-2 & MRPC & STS-B & QQP & MNLI & QNLI & RTE & WNLI  \\
    \toprule
    BERT\textsubscript{BASE} & 57.82 & 92.09 & 86.74 & 88.13 & 87.49 & 84.01 & 90.79 & 64.98 & 53.52\\
    BioBERT & 37.78 & 89.68 & 88.44 & 87.40 & 86.96 & 83.19 & 89.79 & 60.29 & 28.17 \\
    \midrule
    Delta & 20.04 & 2.41 & -1.69 & 0.73 & 0.53 & 0.82 & 1.01 & 4.69 & 25.35  \\
    \bottomrule
    \end{tabular}

    \caption{Performance drop of BioBERT after further pre-training on Pubmed articles.  The last row shows a positive value indicating the degree to which performance has dropped, and a negative value when it has increased.}
    \label{table:motivation}
\end{table*}

\section{Continual Learning}

Our work focuses on three forms of mitigation from catastrophic forgetting.
We explore using constraint based training in the form of EWC, learning rate control from \citet{howard2018universal}, and episodic memory in the form of experience replay.

\subsection{Elastic Weight Consolidation}

EWC makes use of a simple Bayesian factorization of model representation \cite{kirkpatrick2017overcoming}.
This isolates the posterior of a learned task (A) while maintaining the objective of a current task (B).
Due to the intractability of the true posterior, EWC makes use of a Fisher information \citep{frieden2004science} matrix diagonal to approximate the effect of task A on the parameters of a model.
Intuitively speaking, if a parameter had a large effect on task A the Fisher value would be small yielding low variance to adapt to task B.
This holds true inversely for when the Fisher value is large.


In practice, we initialize the Fisher matrix using gradients calculated with data sampled from Task A, which has already converged.
This is demonstrated in Eq. 1 where $i$ and $j$ index parameters and data samples respectively.

\begin{align}
    F_{i,i} &= \frac{1}{N} \sum_{j=1}^{N} \Big( \frac{\partial \mathcal{L}_A^{(j)}}{\partial \theta_i} \Big)^2 \\
    \mathcal{L}(\theta) &= \mathcal{L}_B(\theta) + \sum_i \lambda F_{i,i}(\theta_i - \theta_{A,i}^*)^2
\end{align}
The full objective for task B is given in Eq. 2 where $\mathcal{L}_B(\theta)$ is the objective of Task B, and EWC is represented as the second term regularizing model parameters by weighting the shift of model parameters as it trains on task B ($\theta_i$ and $\theta_{A,i}^*$ being the currently updated and frozen task A parameters at index $i$ respectively).
The EWC objective component is further adjusted by the hyperparameter $\lambda$.

\subsection{Learning rate control}
\label{section:LRC}

Our approach models the second stage of ULMFiT \cite{howard2018universal}, namely target task fine-tuning.
We begin with a layer wise modifications by applying a decaying learning rate as a function of layer depth moving from the last layer towards model input, where $\eta^{(l-1)} = \eta^{(l)}/ 2.6$ ($\eta$ and $l$ denoting learning rate and layer index respectively).
Depth plays a factor in our model since the network consists of 14 layers (i.e. 12 transformer layers, one layer for input, and one for LM heads).
Additionally, we switch from the polynomial decay learning rate scheduler to slanted triangular learning rate (STLR).

\begin{table*}[ht]
    \centering
    \begin{tabular}{lrrr|rr|rr}
    \toprule
    &  \multicolumn{3}{c|}{generic}   &  \multicolumn{2}{c|}{biomedical}  &  \multicolumn{2}{c}{\textit{perplexity}}\\
    Model & GLUE & CoNLL  & MATRES & BC5CDR & Chemprot & BM & nBM  \\
    \midrule
    RoBERTa \textsubscript{BASE} & 87.56 & 90.11 & 79.61 & 84.94 & 63.27 &  4.28 & 1.89\\
    PMC & 83.00 & 87.35 & 71.48 & 86.68 & 65.13 & 2.70 & 7.66 \\
    \midrule
    MDL & 84.89 & 89.72 & 75.37 & 85.76 & 65.16 & 2.91 & 3.95\\
    PMC +LRC & 87.55 & 89.86 & 79.79 & 84.70 & 65.82 & 3.09 & 3.97\\
    PMC +ER & 83.76 & 89.33 & 74.47 & 86.15 & 65.49 & 2.85 & 6.12 \\
    PMC +EWC & 86.27 & 89.91 & 78.78 & 86.11 & 65.47 & 2.83 & 5.14\\
    \bottomrule
    \end{tabular}
    \caption{For each model we report biomedical (BM) and non-biomedical (nBM) perplexity for language modeling, the average accuracy of GLUE, and CoNLL, MATRES, BC5CDR, and Chemprot using micro-$F_1$.}
    \label{table:results_main}
\end{table*}

\subsection{Experience Replay}
We explore experience replay in a very simple fashion.
At a chosen interval we replay a buffer of batches retained from the domain(s) of the previous task.
We explore the frequency of replay as well as the size of the replay data.

\section{Datasets}
We processed publicly available biomedical and non-biomedical corpora for pre-training our models.
For non-biomedical data, we use BookCorpus and English Wikipedia data, CommonCrawl Stories \cite{trinh2018simple}, and OpenWebText \cite{Gokaslan2019OpenWeb}.
This combined corpus contains roughly 18B tokens.
For biomedical data, we use full Pubmed\footnote{https://www.ncbi.nlm.nih.gov/pmc/} articles which we processed to remove all tables, references, equations, and figures.
This yields a dataset of over 4B tokens.
For all datasets we retain training, validation, and test splits sampled at the document level with a respective ratio of 8:1:1.

\section{Experimental Details}
\label{section:details}
For modeling we use the RoBERTa architecture \cite{liu2019roberta}, and implement EWC, learning rate control, and experience replay changes directly into the model.
This extension of the original BERT removed next sentence prediction and trained using only masked language modeling using very large batch sizes.
We utilize all training hyperparameters as provided by \citet{liu2019roberta}
unless otherwise noted, and use RoBERTa \textsubscript{BASE} as parameter initialization for all experiments.

As a form of deterioration understanding, we continue to train a model using Pubmed articles (denoted as PMC) with no mitigation techniques.
For a baseline and potential upper bound of performance we train a multi-domain learning (MDL) model which utilizes the full combined training sets as input data.
The learning rate control model (+LRC) uses the hyperparameters provided by \citet{howard2018universal} and learning rate layerwise decay as outlined in section \ref{section:LRC}.

For EWC (+EWC) we tune both $\lambda$ [$0.5$, \underline{$1.0$}, $5.0$, $10.0$], and the size of the data used for fisher initialization [\underline{$0.1\%$}, $1.0\%$, $10.0\%$]; best values are underlined.
For experience replay (+ER) we experiment with sampling update batches from the non-biomedical dataset (the subset used for EWC init.) at various intervals.
Ten original domain updates at every 1k, 2k, and 5k training steps where each batch size is 2048; a single update of $0.1\%$ of the original domain at the end of an epoch of training.
Best performance was obtained using the latter.

\subsection{Evaluation Data}
To evaluate modeling we track the perplexity of held-out test data for both domains.
We report the average accuracy across GLUE tasks to track the performance of the model on general natural language understanding.
Additionally we evaluate on CoNLL-03 \cite{sang2003introduction} named entity recognition (NER), and MATRES \cite{ning2018multi}, a temporal relation dataset.
To demonstrate domain shift we evaluate using BC5CDR \cite{li2016biocreative} and Chemprot \cite{krallinger2017overview} which are NER and relation extraction (RE) tasks respectively.
The former dataset is from the 2015 CDR challenge for identifying chemicals and diseases expertly annotated from Pubmed abstracts.
Chemprot contains annotations of chemical-protein reactions, also taken from Pubmed articles.

\section{Results}
Our experimental results are highlighted in Table \ref{table:results_main}.
The first two rows contain the off-the-shelf RoBERTa model as well as that which received no mitigation when further trained on biomedical data.
The bottom section lists all other experimental settings described in Section \ref{section:details}.
For all models pre-trained using Pubmed data we finetune on tasks after a single epoch of pre-training.

We divide columns by task domain.
The first three tasks cover general language understanding.
For measuring performance on GLUE, we further limit the selection of tasks to be the five most deteriorated (i.e. CoLA \cite{cola}, SST-2 \cite{sst}, MNLI \cite{mnli}, QNLI \cite{qnli} and RTE \cite{rte}).
Tasks such as QQP\footnote{https://www.quora.com/q/quoradata/First-Quora-Dataset-Release-Question-Pairs} and MRPC \cite{mrpc} are generally robust against domain change and perform well regardless of initialization.
Biomedical tasks are displayed next followed by model perplexity in both domains.
Task scores, save for GLUE, are reported using micro-$F_1$.

\subsection{Catastrophic Forgetting}
We complement our own findings with those from existing pre-trained models.
To this end we fine tuned a BERT\textsubscript{BASE} architecture on all nine GLUE tasks.
These were compared directly against BioBERT, which has been further trained on full Pubmed articles.
Taking a look at Table \ref{table:motivation} an overall trend of performance deterioration is apparent with a relative increased error of $7.64 \%$.
BioBERT performed negligibly better than original BERT on only a single task (MRPC).
Furthermore, we observed that on tasks which BERT struggles with, such as CoLA and WNLI, the performance decrease is amplified when switching pre-training domains.

Our own results are similarly divided.
Unsurprisingly among these RoBERTa \textsubscript{BASE} performs best on GLUE, CoNLL and MATRES.
Conversely it under-performs on the biomedical tasks, validating the need to further pre-train on domain specific data.
Similarly we see that the PMC model performs the best in its domain however there is significant drop in performance across GLUE, CoNLL amd MATRES.
The perplexity analysis further illustrates the degree of separation between tasks, with the biomedical model exhibiting a sharp change when leaving the generic domain.

\subsection{Mitigation based models}
EWC and LRC both respond well during domain shifts and are our best candidates for combating catastrophic forgetting.
LRC has negligible degradation on GLUE tasks, and yields best overall numbers for MATRES, and Chemprot.
Furthermore this model has the highest combined confidence when observing perplexity across domains.
Our trials with EWC left us with several findings.
While the amount of data used for Fisher initialization did not have a profound effect, the model was quite sensitive to $\lambda$ values.
With higher coefficients ($\lambda > 1.0$) EWC was able to halt deterioration nearly completely but performed quite poorly on biomedical tasks.
To better understand the importance of fisher, we trained EWC with no Fisher (i.e removing $F_{i,i}$ from Eq. 2). We found that this resulted in lower biomedical results, which shows that giving equal weight to all the parameters results in poor generalization on source and target domains.
MDL performed surprisingly average compared to the resource trade-off of the model.
While it does produce competitive results in the biomedical domain, the model struggles to retain generic knowledge.
Experience replay grapples most with domain retention and produced the highest mitigated biomedical results coupled with the lower generic results.

\subsubsection{Two stage domain shift}
To further evaluate the robustness of the best performing methods we add a third domain to the continual learning setup.
We processed 659M tokens of de-identified clinical notes and continued training the EWC, and LRC from Table \ref{table:results_main} (denoted with a subscript 2).
Evaluating the clinical domain we use NER from the 2010 i2b2 challenge.
Due to the relatively small amount of clinical data we pre-train for five epochs.
We compare against the deterioration on an unmitigated model trained first on Pubmed, and then clinical data (PMC, clin.).
\begin{table}[h]
    \centering
    \begin{small}
    \begin{tabular}{lrrrr}
    \toprule
    Model & GLUE & CoNLL & BC5CDR & i2b2\\
    \midrule
        RoBERTa \textsubscript{BASE} & 87.56 & 90.11 & 84.94 & 81.12\\
        PMC, clin. & 80.08 & 86.18 & 85.05 & 84.74\\
        LRC\textsubscript{2} & 87.99 & 89.67 & 84.54 & 82.66\\
        EWC\textsubscript{2} & 85.06 & 88.32 & 86.00 & 85.26\\
    \bottomrule
    \end{tabular}
    \caption{GLUE and NER values for models trained across three domains: generic, biomedical, and clinical.}
    \end{small}
    \label{table:two_stage}
    
\end{table}

As expected the unmitigated model suffers from performance deterioration in both previous domains, with average GLUE dropping drastically.
LRC worsens below baseline on BC5CDR and shows only a small boost in clinical results over RoBERTa \textsubscript{BASE}, although it continues to perform well on generic tasks.
EWC gives the best performance across the board.
The model exhibits slight decay on GLUE and CoNLL, robust performance on biomedical NER, and the best overall results on i2b2.
This further indicates that the EWC objective has the capability to produce a model which generalizes better across multiple domains, outperforming unregulated methods.

\section{Conclusion}

In this work, we have demonstrated the existence of catastrophic forgetting in large language model pre-training.
We further explored constraint and replay based mitigation techniques to close the performance gap between general and domain specific natural language tasks.

\bibliography{main}
\bibliographystyle{acl_natbib}

\end{document}